\documentclass[runningheads]{llncs}

\usepackage[T1]{fontenc}
\usepackage{graphicx}

\usepackage{algorithm}
\usepackage{algpseudocode}
\usepackage{amsmath}
\usepackage{booktabs}
\usepackage{multirow}
\usepackage{caption}
\usepackage{float}
\usepackage{graphicx}
\usepackage[caption=false]{subfig}
\begin{document}
\title{A Quantum-Driven Evolutionary Framework for Solving High-Dimensional Sharpe Ratio Portfolio Optimization}
\titlerunning{Quantum-Driven DE for High-Dimensional Portfolio Optimization}
%
\author{Mingyang Yu$^{\dagger}$\inst{1}\orcidID{0009-0000-5903-6014} \and
Jiaqi Zhang$^{\dagger}$\inst{1}\orcidID{0009-0008-1306-3146} \and
Haorui Yang\inst{1}\orcidID{0009-0001-5009-4872} \and
Adam Slowik\inst{2}\orcidID{0000-0003-2542-9842} \and
Jun Zhang\inst{1, 3}\orcidID{0000-0001-7835-9871} \and
Jing Xu$^{*}$\inst{1}\orcidID{0000-0001-8532-2241}
}
\authorrunning{M. Yu et al.}
%
\institute{College of Artificial Intelligence, Nankai University, Tianjin 300350, China. 
	\email{1120240312@mail.nankai.edu.cn, 2120250616@mail.nankai.edu.cn, 2120240645@mail.nankai.edu.cn, xujing@nankai.edu.cn}
	\and
Department of Electronics and Computer Science, Koszalin University of Technology, Koszalin 75-453, Poland \email{adam.slowik@tu.koszalin.pl} 
\and
Hanyang University ERICA Campus, South Korea \email{junzhang@ieee.org}
\\
$^\dagger$Equal contribution. $^*$Corresponding authors}

\maketitle              
\begin{abstract}
High-dimensional portfolio optimization faces significant computational challenges under complex constraints, with traditional optimization methods struggling to balance convergence speed and global exploration capability. To address this, firstly, we introduce an enhanced Sharpe ratio-based model that incorporates all constraints into the objective function using adaptive penalty terms, transforming the original constrained problem into an unconstrained single-objective formulation. This approach preserves financial interpretability while simplifying algorithmic implementation. 
To efficiently solve the resulting high-dimensional optimization problem, we develop a Quantum Hybrid Differential Evolution (QHDE) algorithm, which introduces a dynamic quantum tunneling mechanism that enables individuals to probabilistically escape local optima, dramatically enhancing global exploration and solution flexibility. To further improve performance, a good point set–chaos reverse learning strategy generates a well-dispersed initial population, providing a robust and diverse starting point. Meanwhile, a dynamic elite pool combined with Cauchy-Gaussian hybrid perturbations maintains population diversity and mitigates premature convergence, ensuring stable and high-quality solutions.
Experimental validation on CEC benchmarks and real-world portfolios involving 20 to 80 assets demonstrates that QHDE's performance improves by up to 96.6\%. It attains faster convergence, higher solution precision, and greater robustness than seven state-of-the-art counterparts, thereby confirming its suitability for complex, high-dimensional portfolio optimization.

\end{abstract}
\section{Introduction}
In finance, a portfolio represents a collection of stocks or assets managed by individuals or institutions. Portfolio optimization entails distributing a fixed budget among the assets to minimize risk while maximizing expected returns. The Modern Portfolio Theory (MPT) employs the mean-variance model to describe the relationship between risk and return \cite{1}. The core principle of this theory quantifies risk using variance, where lower-risk portfolios are typically characterized by smaller variances. However, MPT overlooks practical market constraints and the influence of background risks, such as base requirements, upper and lower bounds, and challenges inherent in managing large-scale asset portfolios \cite{2,3,4}. Therefore, real-world portfolio optimization must incorporate these constraints and uncertainties to account for actual market conditions.

Differential Evolution (DE) \cite{18}, as a classic heuristic optimization algorithm, is widely applied in various engineering optimization and scientific research tasks due to its simple implementation, few parameters, and strong global search capability \cite{25,ZHONG2025112777,ZHANG2026112587}. Given the hybrid decision-making nature of discrete asset selection and continuous weight allocation in portfolio optimization, the real-valued encoding mechanism of DE facilitates a precise characterization of the continuous search space of asset weights, and combined with specialized mapping strategies to effectively handle complex constraints and multi-objective optimization, thereby adapting to both nonlinear and multi-modal requirements in financial decision-making. However, when confronted with high-dimensional complex portfolio problems, DE faces notable limitations. First, it is prone to becoming trapped in local optima, hindering the search for the global optimum. Second, as the dimensionality increases, the search efficiency declines and convergence slows. These limitations restrict the application effectiveness of DE in the field of portfolio optimization. 

To overcome the aforementioned shortcomings, researchers have attempted to enhance the population diversity of the algorithm by introducing quantum-inspired mechanisms such as wave functions, probability amplitudes, and qubit representations \cite{YU2026103334,a1}. For instance, the evolution of probability density driven by the Schr\"{o}dinger equation \cite{sun2004particle}, operator reconstruction based on qubit representations \cite{agrawal2020quantum}, and the probability amplitude directional collapse induced by the quantum rotation gate \cite{yu2026alzheimer} have been proven to significantly improve the global exploration efficiency of the algorithm. However, existing quantum-inspired operators often lack dynamic state awareness, leading to poor search precision in later iterations. Additionally, they struggle to balance energy barrier penetration with local exploitation when facing objective functions with complex rotation and shift characteristics. 

In response to the above challenges, we propose a variant of the DE based quantum theory, referred to as the Quantum Hybrid Differential Evolution Algorithm (QHDE). QHDE incorporates multiple innovative improvements relative to conventional DE. Inspired by the quantum tunneling behavior, QHDE constructed a dynamic quantum potential well model based on fitness differences. This enables individuals to bypass the local traps that were traditionally difficult for classical DE to overcome. Furthermore, the incorporation of chaotic reverse learning extends the coverage of initial solutions, resulting in a more advantageous search starting point. Meanwhile, the dynamic elite pool and Cauchy-Gaussian mixed disturbance strategy effectively prevent premature convergence and increase the diversity of the solutions. The combination of these improvement strategies enables QHDE to surpass traditional DE and other metaheuristic algorithms in portfolio optimization, allowing for faster identification of the optimal solution that balances risk and return. The main contributions of this paper are as follows:

(1)	We develop a quantum tunneling strategy that transcends classical search limitations. By enabling non-classical probabilistic transitions, this strategy allows individuals to bypass local optima through a 'tunneling' effect rather than stochastic jumps, fundamentally enhancing global exploration efficiency.

(2)	We introduce an initialization mechanism combining the good point set method with chaotic reverse learning. This enhances the quality of initial solutions, offering a more efficient starting point for subsequent optimization.

(3)	We propose a dynamic elite pool strategy and applies Cauchy-Gaussian random perturbation. This strategy mitigates premature convergence to local optima, enhances solution diversity, and strengthens global search capability.

(4)	We apply QHDE to portfolio optimization, where it outperforms seven state-of-the-art algorithms, exceeding the runner-up by up to 96.6\%. This approach significantly accelerates convergence while enhancing accuracy, proving to be an efficient and robust solution for high-dimensional problems.

\section{Math Equations}
This section introduces modeling methods of the portfolio selection problem, laying the foundation for the subsequent introduction of optimization techniques.

\subsection{Classical Models}
\textbf{Mean-Variance Model:} 
The objective function minimizes total portfolio risk, treating the expected return as a constraint. This ensures the portfolio meets a specific return threshold while maintaining the lowest possible risk level:

\begin{equation}
	\min \sum_{i=1}^{M} \sum_{j=1}^{M} E_i E_j Q_{ij}
\end{equation}

\begin{equation}
	\text{s.t.} \sum_{i=1}^{M} E_i \alpha_i = R
\end{equation}
where, $M$ denotes the total number of stocks, $E_i$  and $E_j$ represents the investment weight of the $i$th and $j$th stock respectively, $Q_{ij}$ signifies the covariance between the $i$th and $j$th stocks, and $\alpha_i$ denotes the expected return of the $i$th stock. 
Equation (2) ensures that the total return of the investment portfolio reaches the predetermined expected return $R$. 

\textbf{Efficient Frontier:} 
Unlike the mean-variance model, which treats expected returns as a constraint, the efficient frontier model integrates them directly into the objective function. It optimizes the risk-return trade-off by applying a penalty term to the return, and the expression is:
\begin{equation}
	\min \left[ \omega \left( \sum_{i=1}^{M} \sum_{j=1}^{M} E_i E_j Q_{ij} \right) 
	- (1 - \omega) \left( \sum_{i=1}^{M} E_i \alpha_i \right) \right]
\end{equation}
where $\omega \in [0,1]$. 

\textbf{Sharpe Ratio Model:} 
The Sharpe Ratio Model shifts focus from individual parameters to their ratio, measuring excess return per unit of risk. This single metric comprehensively balances portfolio performance and risk, expressed as:

\begin{equation}
	Sharpe- ratio = \frac{R_p - R}{std(p)}
\end{equation}
where $p$ denotes a portfolio, $R_p$ is the return of the portfolio, and $R$ is the risk-free rate of return. $std(p)$ represents the standard deviation of the portfolio, which can be considered as the risk of portfolio. 
By specifying $R_p$ and $std(p)$ in Equation (4) and integrating the previously mentioned constraints, the complete optimization model is given as:
\begin{equation}
	\max \frac{\sum_{i=1}^{M} E_i \alpha_i - R}{\sqrt{\sum_{i=1}^{M} \sum_{j=1}^{M} E_i E_j Q_{ij}}}
\end{equation}

In all the above models, the investment weights $E_i$ are subject to full allocation ($\sum E_i = 1$) and non-negativity ($0 \le E_i \le 1$) constraints. This ensures that the funds are fully allocated and no short-selling is allowed.

\subsection{Our unconstrained model}
In this study, we refer to the Sharpe ratio model and reference \cite{2} to transform the original constrained optimization model into an unconstrained single-objective optimization problem. The transformed objective function is:
\begin{equation}
	\begin{aligned}
		F(E) = \max &\; \beta_1 \left[ \frac{\sum_{i=1}^{M} E_i \alpha_i - R}{\sqrt{\sum_{i=1}^{M} \sum_{j=1}^{M} E_i E_j Q_{ij}}} \right] 
		- \beta_2 \left( \sum_{i=1}^{M} E_i - 1 \right)^2 \\
		& - \beta_3 (E_i - 1 \geq 0)(E_i - 1) 
		- \beta_4 (-E_i \geq 0)(-E_i)
	\end{aligned}
\end{equation}
where $\left( \sum_{i=1}^{M} E_i - 1 \right)^2$ represents the equality constraint specified, ensuring the sum of the investment weights equals 1. While $(E_i - 1 \geq 0)(E_i - 1)$ and $(-E_i \geq 0)(-E_i)$ correspond to the inequality constraints outlined, which ensure the weights remain within valid bounds. $\beta_1$, $\beta_2$, $\beta_3$, and $\beta_4$  denote the penalty weights associated with these constraints. 

\section{Methodology Overview: original DE and Proposed Improvements}
In this section, we first detail the original DE, then introduce the proposed QHDE, focusing on: a good point set-chaos reverse learning initialization, a potential-driven dynamic quantum tunneling strategy, and a dynamic elite pool with Cauchy-Gaussian mixed disturbance.

\subsection{The original DE}
DE is a classic population-based intelligent optimization algorithm, widely used in global optimization problems in continuous spaces. Its core consists of four operations: initialization, mutation, crossover, and selection. In each generation $t$, for individual $x_r (t)$, the mutation vector is generated as:
\begin{equation}
	v_i(t+1) = x_{r_1}(t) + F \cdot (x_{r_2}(t) - x_{r_3}(t))
\end{equation}
where $x_{r_1}$, $x_{r_2}$ and $x_{r_3}$ denote different individuals randomly selected from the population; $F$ represents the mutation parameter, typically taking values in the range [0,2]; and $t$ represents the iteration count.
Subsequently, a binomial crossover operation is employed to generate the trial vector $u_{i,j}$
\begin{equation}
	u_{i,j}(t+1) =
	\begin{cases}
		v_{i,j}(t+1) & \text{if } \text{rand}_j(0,1) \leq CR \text{ or } j = j_{\text{rand}} \\
		x_{i,j}(t)   & \text{otherwise}
	\end{cases}
\end{equation}
where $u_{i,j}$ represents the $j$th component of the trial vector $u_i$, $rand_j (0,1)$ represents a random number in the range [0,1], $CR$ denotes the crossover probability, and $j_{\text{rand}}$ denotes a randomly selected index to ensure that $u_i$ has at least one component from $v_i$.
Finally, if the fitness of the test individual is superior to that of the original individual, it will enter the next generation:
\begin{equation}
	x_i(t+1) =
	\begin{cases}
		u_i(t+1) & \text{if } f(u_i(t+1)) \leq f(x_i(t)) \\
		x_i(t)   & \text{otherwise}
	\end{cases}
\end{equation}
where $f(x)$ denotes the fitness value of individual $x$.

\subsection{Our proposed QHDE}

Portfolio optimization in high-dimensional spaces presents significant challenges due to nonlinear constraints and numerous local optima. To address these, we enhance DE by integrating quantum tunneling, adaptive mechanisms, and efficient perturbation strategies. These improvements strengthen global search capabilities and mitigate premature convergence. Consequently, the proposed algorithm demonstrates superior robustness and precision, effectively navigating complex search spaces to achieve stable, high-quality solutions for large-scale portfolio problems.

\subsubsection{A good point set-chaos reverse learning initialization mechanism}

In metaheuristic algorithms, the quality of the initial solution significantly influences the search outcome. Standard pseudo-random generators often produce uneven distributions or clusters. Hence, to ensure maximum uniformity and ergodicity across the search space, we utilize the good points set to initialize the population. 
In the $s$-dimensional unit hypercube $G_s$, the set of points for $r \in G_s$ is defined as:

\begin{equation}
	GP_n(k) = \left( \left\{ r_1^{(n)} \times k \right\}, \ldots, \left\{ r_s^{(n)} \times k \right\} \right), \quad 1 \leq k \leq n
\end{equation}

If the deviation satisfies $\varphi(n) = C(r,\varepsilon)n^{-1+\varepsilon}$, where $C(r,\varepsilon)$ represents a constant that depends only on $r$ and $\varepsilon$, $GP_n(k)$ is referred to as the good point set. Let $r = \{2\cos(2\pi i/pn), 1 \leq i \leq s\}$, where $pn$ represents the smallest prime number satisfying $(p - 3)/2 \geq s$, and $r$ represents termed a good point. After generating the good point set, it is mapped to the search space as:

\begin{equation}
	X_{i,j} = (ub - lb) \times \{GP_n(k)\} + lb
\end{equation}
where $X$ represents the initial solution, $X_{i,j}$ denotes the $j$th component of the $i$th individual, and $ub$ and $lb$ respectively represent the upper and lower bounds of the search space.

Reverse learning is a perturbation mechanism that compares a known solution with its reverse counterpart in the target space, selecting the superior one. Chaotic systems, known for their sensitivity, ergodicity, and unpredictability, provide a more robust source of randomness than traditional sequences. In this paper, we adopt the Logistic chaotic mapping to initialize the population X generated by the optimal point set:
\begin{equation}
	G^{t+1} = \mu \times G^t (1 - G^t)
\end{equation}
where, $\mu$ represents the chaotic coefficient, $\mu \in (2,4)$, and $G$ represents the chaotic factor generated by the Logistic chaotic mapping. The initialization process combines the Logistic chaotic mapping and reverse learning as follows:
\begin{equation}
	X' = G \times (ub - lb) - X
\end{equation}
where $X'$ denotes the solution adjusted through reverse learning. After merging $X'$ and $X$, $N$ individuals with the top one-half fitness values are selected as the new population.

\subsubsection{Potential-driven dynamic quantum tunneling strategy}
Traditional DE often suffers from slow convergence and local optima stagnation, especially in complex tasks like portfolio optimization.
To address these challenges, this paper proposes integrating quantum tunneling theory into the DE framework. This mechanism models a individuals trapped in local optima as a particle confined within a time-variant potential well, defined by center $C$ and characteristic length $L$. As iterations $t$ progress, the potential well expands, enabling a non-classical probabilistic transition. Guided by the tunneling probability $P_t(X_i)$, solutions can "tunnel" through energy barriers (fitness gaps) to relocate beyond local basins. This probabilistic mechanism significantly boosts global exploration and the algorithm's ability to reach the global optimum. The strategy is structured into three primary components:

\textbf{1. Potential Energy Modeling:} 
The fitness difference of an individual $x_i$ is mapped to a quantum potential barrier $V(x_i)$, defined as follows:
\begin{equation}
	V(x_i) = \frac{|fit_i - f_{\text{best}}|}{\max(f) - \min(f)} \cdot V_0
\end{equation}
where $f_{\text{best}}$ denotes the current global best fitness, and $V_0$ denotes the baseline barrier height. This model ensures that individuals closer to the optimum face lower potential barriers, while those further away encounter higher resistance.

\textbf{2. Tunneling Probability Calculation:} 
Based on the Wentzel-Kramers-Brillouin (WKB) approximation, the equivalent tunneling probability $P_t(x_i)$ is formulated to determine the likelihood of an individual penetrating the "energy barrier":
\begin{equation}
	P_t(x_i) = \exp\left( -\frac{2\sqrt{2m}}{\hbar} \int_{a}^{b} \sqrt{V(x) - E_k} \, dx \right)
\end{equation}
where $E_k = \frac{1}{N} \sum_{i=1}^{N} f(x_i)$ represents the average kinetic energy of the population, $\gamma \in [0.05, 0.2]$ denotes the tunneling adjustment coefficient. When the potential energy $V(x_i)$ is lower than $E_k$, $P_t$ is set to 1, facilitating free movement. Conversely, if $V(x_i) > E_k$, the probability decreases exponentially, allowing for controlled, non-classical probabilistic jumps.

\textbf{3. Intelligent Position Update:} 
If the tunneling condition ($rand < P_t(x_i)$) is triggered, the individual's position is updated via a coupled perturbation mechanism:
\begin{equation}
	x_{\text{new}} = x_i + \beta \cdot T(t) \cdot \Delta(t)
\end{equation}
In this update rule, $\beta \sim \mathcal{N}(0,1)$ follows the standard normal distribution, providing a random search direction. While the time-varying coefficient $T(t)$ and the dynamic step size $\Delta(t)$ collaboratively balance global exploration in early stages with local fine-tuning in later iterations.

\subsubsection{Dynamic elite pool with Cauchy-Gaussian mixed disturbance strategy}
In later iterations, DE populations often cluster around the current best solution. This rapid assimilation causes stagnation and premature convergence if that solution is a local optimum. To counter this, we propose a dynamic elite pool ($Elite_p$) combined with a Cauchy-Gaussian mixed disturbance strategy. By maintaining a pool of top-performing individuals and applying stochastic perturbations, the algorithm preserves population diversity and strengthens its global search capacity. The specific procedure is as follows:

\begin{figure*}[t]
	\centering
	\includegraphics[width=\textwidth]{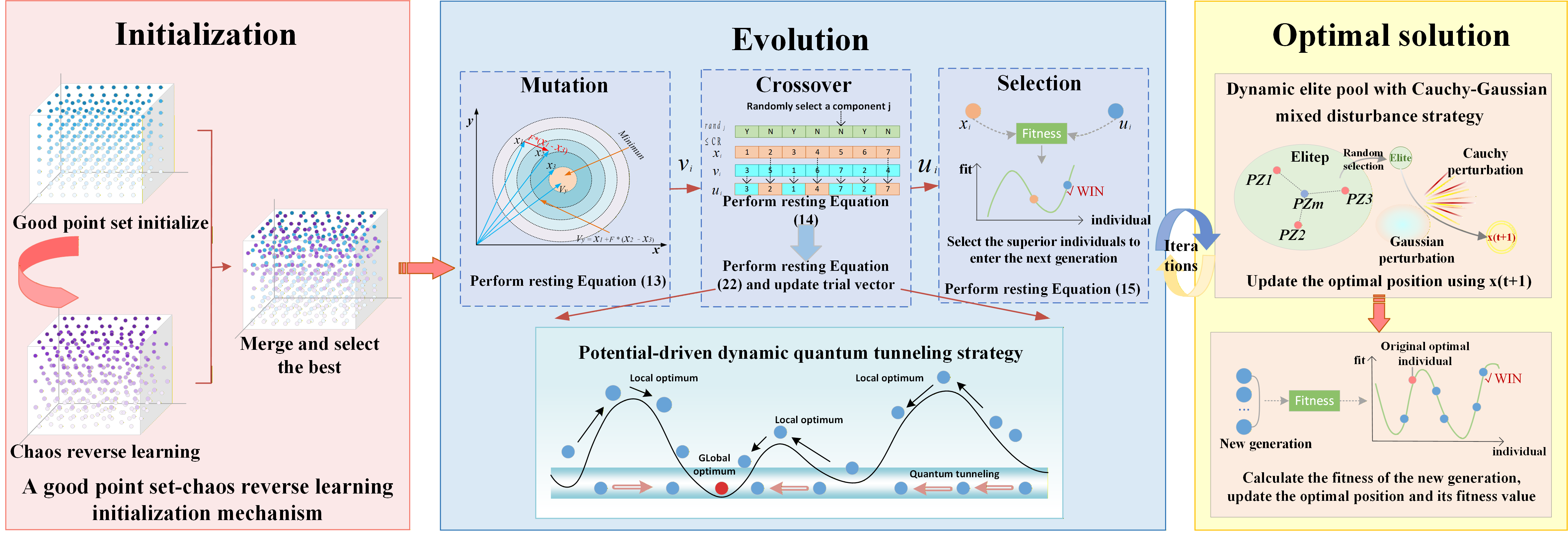}%
	\caption{Flow chart of QHDE}
	\label{fig:QHDE}
\end{figure*}

First, the three highest fitness individuals, denoted as $PfZ1$, $PZ2$, and $PZ3$, are selected from the population. Then the arithmetic mean of their positions is calculated to obtain the average position $PZm$:

\begin{equation}
	PZm = \frac{(PZ1 + PZ2 + PZ3)}{3}
\end{equation}

Next, these three top individuals, along with their average position, are included in \textit{Elitep}. During each iteration, a position is randomly selected from \textit{Elitep} as a reference point to guide the movement direction of individuals.

Cauchy-Gaussian mutation is a hybrid mutation method that combines the long-tail characteristics of the Cauchy distribution with the concentration tendency of the Gaussian distribution. The purpose of this strategy is to apply strategic random perturbations to the global best individual, exploring potentially unknown beneficial regions around the global optimum, while maintaining the "memory" of the current best solution. After combining the dynamic elite pool and Cauchy-Gaussian perturbation strategies, the update formula for the new individual's position is as follows:
\begin{equation}
	x^{t+1}_{\text{best}} = Elite \times [\rho \times \text{Cauchy}(0,1) + (1 - \rho) \times \text{Gauss}(0,1)]
\end{equation}
where $x^{t+1}_{\text{best}}$ represents the global best individual in $t+1$ generation, and $\rho \in [0,1]$ denotes a weight coefficient balancing the Cauchy and Gaussian distributions.

In summary,  the pseudocode and workflow of the QHDE is presented in Algorithm 1 and Figure ~\ref{fig:QHDE}.

\begin{algorithm}[t]
	\caption{Quantum Hybrid Differential Evolution (QHDE)}
	\begin{algorithmic}[1]
		\State \textbf{Initialize} Problem Setting (dimension ($D$), $ub$, $lb$, population ($N$), iterations ($T_{\max}$))
		\State Create an initial population $X = (x_1, x_2, \ldots, x_N), x_i \in D$ using (16), (17)
		\State Generate the population $X'$ using (18), (19)
		\State Merge $X'$ with $X$, and select $N$ individuals with the highest fitness values as the new population $P$
		\While{$t \leq T_{\max}$}
		\For{$i = 1$ to $N$}
		\State Randomly select $x_{r_1}, x_{r_2}, x_{r_3}$ from the population and calculate mutant vector $v_i$ using (13)
		\State Generate trial vector $u_i$ using (14)
		\State Calculate the tunneling probability $P_t(x_i)$ using (21)
		\State Generate $x_{new}$ using (22) 
		\If{$x_{new}$ surpasses $u_i$}
		\State $u_i = x_{new}$
		\EndIf
		\If{$f(u_i) \leq f(x_i)$}
		\State Insert $u_i$ into new generation $P_{new}$
		\Else
		\State Insert $x_i$ into new generation $P_{new}$
		\EndIf
		\EndFor
		\State Select the fittest individuals $PZ1, PZ2, PZ3$ and calculate $PZm$ using (23) to form the elite pool
		\State Generate $x_{\text{best}}^{t+1}$ using (24)
		\If{$x_{\text{best}}^{t+1}$ surpasses $pos_{\text{best}}$}
		\State $pos_{\text{best}} = x_{\text{best}}^{t+1}$
		\EndIf
		\State Update best global position $pos_{\text{best}}$ and its fitness value $fit_{\text{best}}$
		\State $P = P_{new}$
		\State $t = t + 1$
		\EndWhile
		\State \Return $pos_{\text{best}}$ and $fit_{\text{best}}$
	\end{algorithmic}
\end{algorithm}

\subsection{Time and space complexity analysis}
The computational complexity of QHDE mainly depends on population initialization and core operations such as fitness evaluation and iterative updates. With population size $N$, iteration limit $T_{\max}$, and dimension $D$, the original complexity is $O(T_{\max}\times N\times D)$. Initialization operations (good-point-set generation, chaotic mapping, and reverse learning) each cost $O(N\times D)$ and are executed once. Quantum tunneling updates and the elite-pool disturbance strategy both require $O(N\times D)$ per iteration, giving $O(T_{\max}\times N\times D)$. Thus, the total complexity remains $O(T_{\max}\times N\times D)$, consistent with the original algorithm.

The space complexity of DE depends on storing the population, fitness values, and candidate solutions. With population size $N$ and dimension $D$, storing positions and temporary candidates requires $O(N\times D)$, while fitness values need $O(N)$. Thus, the overall complexity is $O(N\times D)$. In the improved QHDE, storing good-point sets, reverse individuals, and the elite pool also requires $O(N\times D)$, so the total space complexity remains $O(N\times D)$.

\section{Experiments and comprehensive analysis}
This section provides a detailed discussion of the experimental results obtained by applying QHDE to numerical optimization tasks and portfolio selection problems. We first use the  CEC 2020 \cite{38} and CEC 2022 \cite{39} suites to test its efficacy on multimodal and high-dimensional functions. Subsequently, QHDE is applied to four portfolio problems (20, 40, 60, and 80 stocks) and compared against seven widely-used metaheuristic algorithms to validate its superiority.
The simulations were conducted on a Windows 11 platform with a 64-bit operating system. The analyses were carried out using MATLAB 2023b on a system powered by an AMD Ryzen 7 4800H CPU at 2.30 GHz and 16 GB of RAM.

\subsection{Test functions and comparison algorithms}
To evaluate QHDE, we utilized the CEC 2020 and 2022 test suites. These benchmarks encompass diverse, high-dimensional, and multimodal optimization problems that effectively mirror the non-convex objectives and complex constraints of real-world portfolio selection. Their emphasis on balancing multiple local optima and risk-return trade-offs provides a rigorous, representative basis for assessing the algorithm's effectiveness and robustness in complex environments.

The performance of QHDE was systematically compared against seven well-known algorithms. These are categorized into two distinct groups for a structured analysis: (1) State-of-the-art algorithm: SASS \cite{10.1145/3377929.3398186}, COLSHADE \cite{gurrola2020colshade}, sCMAES \cite{ros2008simple}, HSEPSO \cite{wei2025hsepso}, NDSOT \cite{fujita2025adaptive}; (2) DE \cite{18} and its improved variants ADE \cite{42}. Table~\ref{tab:params} provides an overview of the parameter settings for eight algorithms. 

\begin{table}[t]
	\centering
	\caption{Parameter configurations for competing algorithms}
	\label{tab:params}
	\setlength{\tabcolsep}{15pt}
	\begin{tabular}{ccl}
		\toprule
		\textbf{Algorithm} & \textbf{Parameter} & \textbf{Value} \\
		\midrule
		
		SASS & $NP_{min}, H, P_{gr}, \sigma$ & 4, 5, 0.2, 0.1 \\
		COLSHADE   & H, r, p, $\alpha$              & 5, 2.6, 0.11, 1.5 \\
		sCMAES  & $\sigma$            &  0.3\\
		HSEPSO   & $w$, $c$                     & [0.4, 0.95], [1.2, 2] \\
		NDSOT & TOL,MAX-INC, $w$, $c_3$, $c_4$      & $10^{-6}$, 10, 0.5, 1.0, 1.0 \\
		DE    & $\beta$, $P_c$                     & 0.5, 0.5 \\
		ADE   & $\beta$, $P_c$                     & [0.2, 0.8], 0.5 \\
		QHDE  & $\beta$, $P_c$, $V_0$, $\gamma$ & 0.5, 0.1, 0.5, 0.1 \\
		\bottomrule
	\end{tabular}
\end{table}

\subsection{Quantitative evaluation}
This section evaluates QHDE's performance using the CEC 2022 test suite (CEC 2020 results detailed in the Appendix \cite{QHDE}). To ensure statistical reliability, we standardized the experimental parameters: a population size of 30, a maximum of 500 iterations, and 30 independent runs for each algorithm.

\begin{figure}[t]
	\centering
	\includegraphics[width=0.5\linewidth]{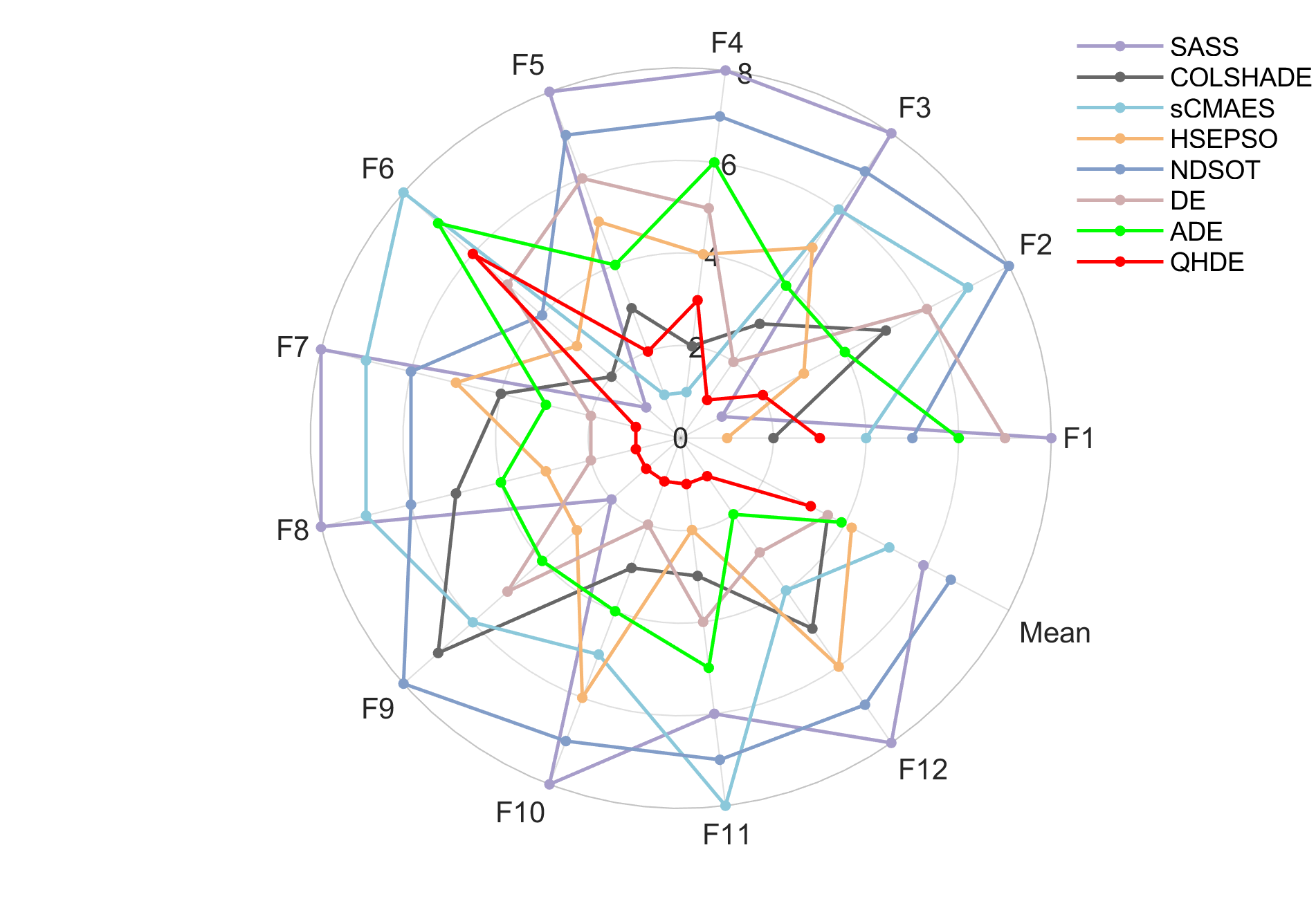}%
	\hfill
	\includegraphics[width=0.5\linewidth]{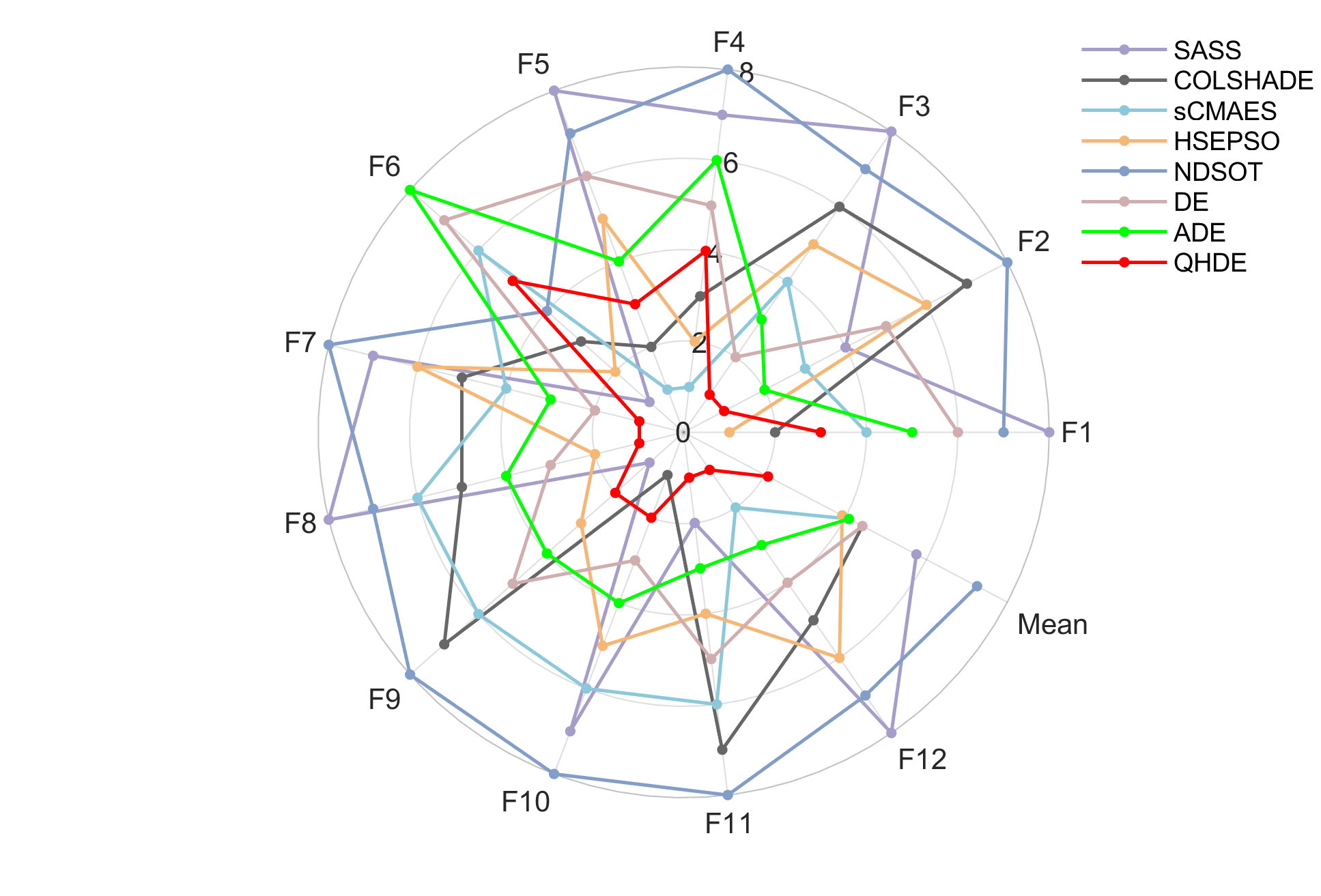}
	\vspace{1pt}
	{\small (a) \hspace{0.5\linewidth} (b)}
	\caption{Ranking distribution of different algorithms on CEC 2022. (a) Dim = 10, (b) Dim = 20.}
	\label{fig:ranking_cec2022}
\end{figure}

Figure~\ref{fig:ranking_cec2022} presents the Friedman rank radar plot of all participating algorithms. As evident in Figure~\ref{fig:ranking_cec2022}(a), QHDE (red) consistently occupies the central region, indicating superior (lower) rankings. As dimensionality increases to 20 in Figure~\ref{fig:ranking_cec2022}(b), although the QHDE curve shows a broader distribution due to heightened problem complexity, it remains significantly more compact than its competitors. Notably, QHDE ranks first on functions F2, F3 and F7-12, which are known for their high complexity and multimodality. The overall rank of QHDE is the lowest among all contenders, further confirming its superior and robust optimization capability across a wide range of scenarios.

\begin{figure}[t]
	\includegraphics[width=0.32\linewidth]{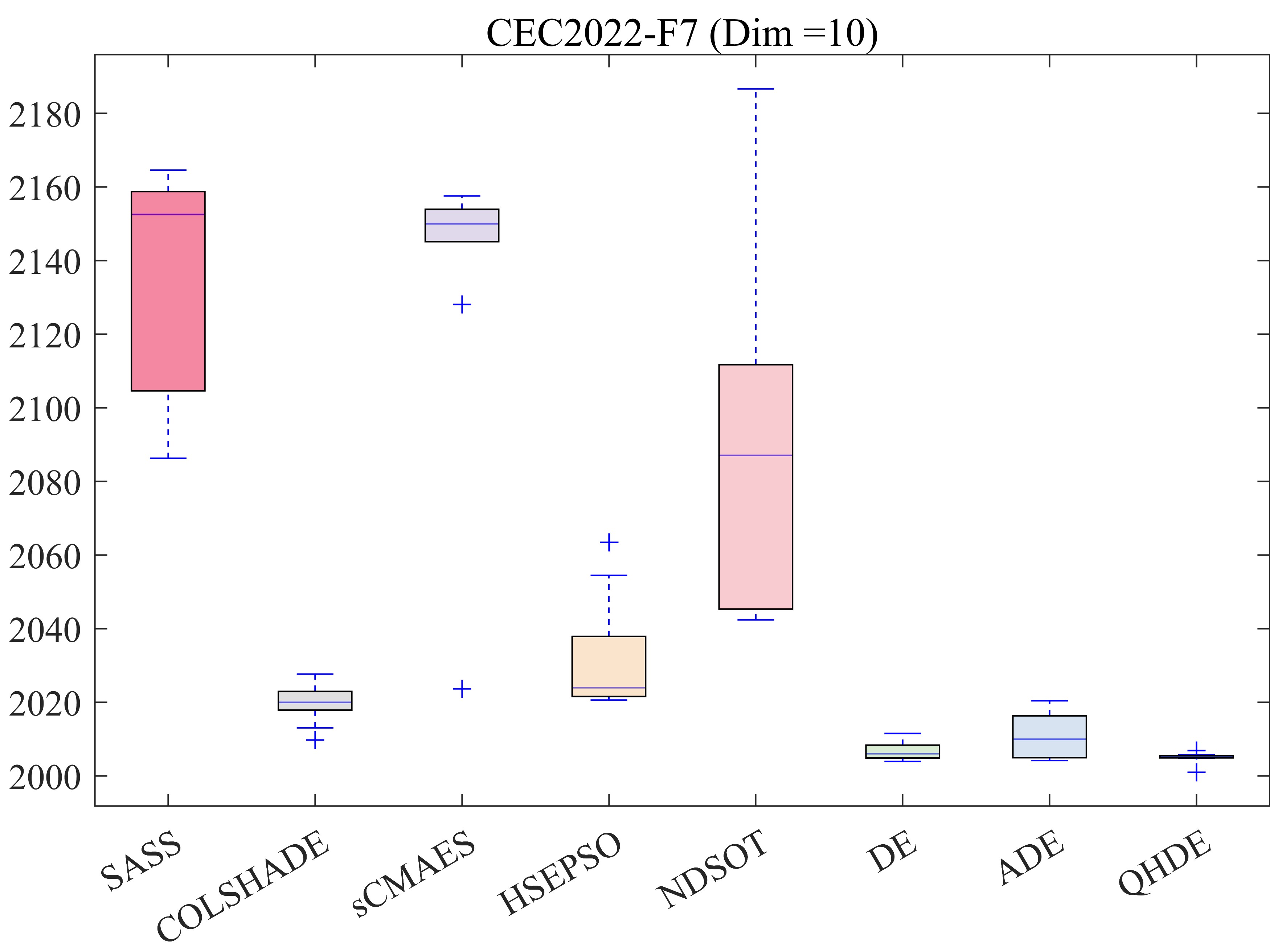}%
	\hfill
	\includegraphics[width=0.32\linewidth]{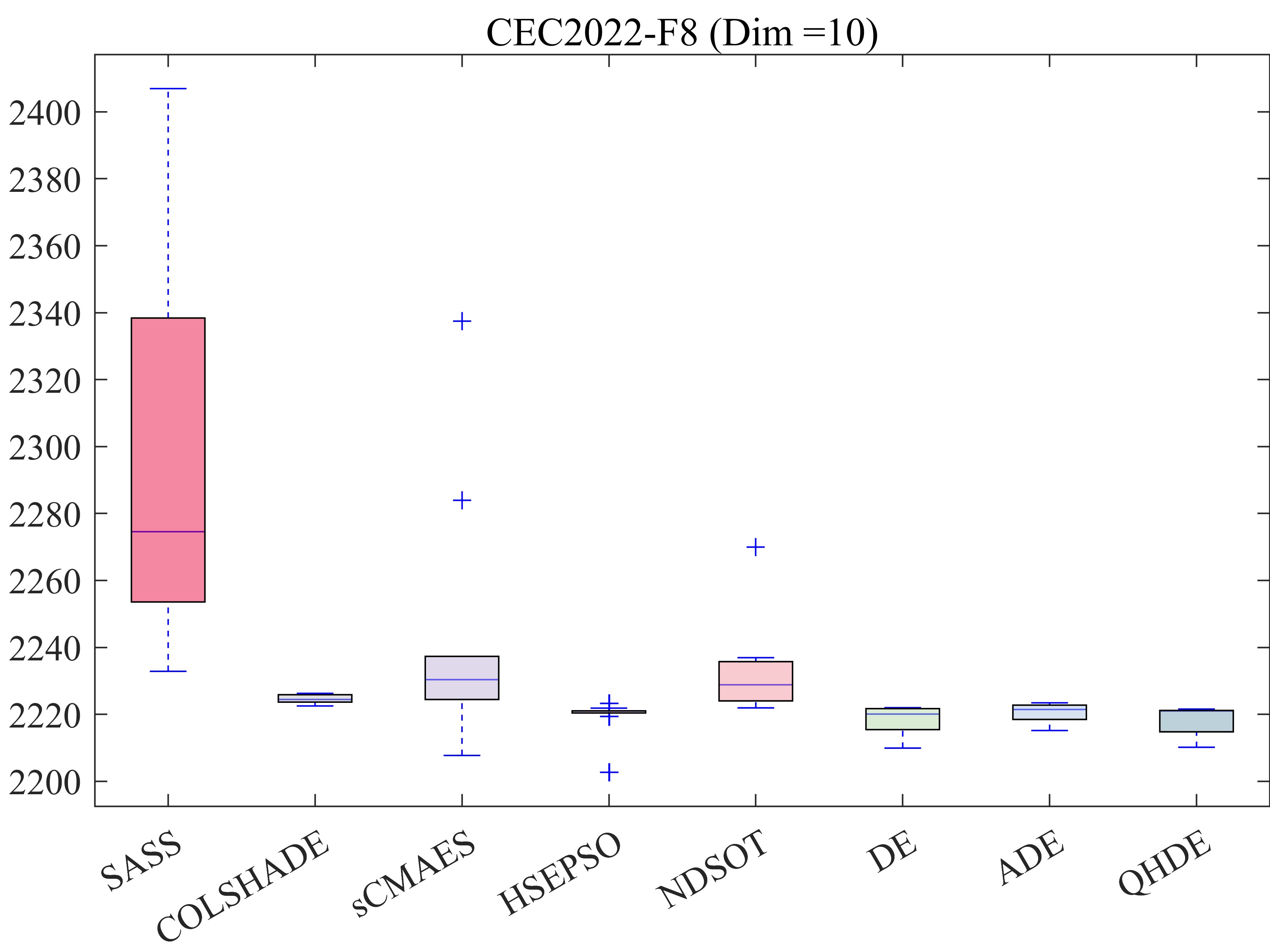}%
	\hfill
	\includegraphics[width=0.32\linewidth]{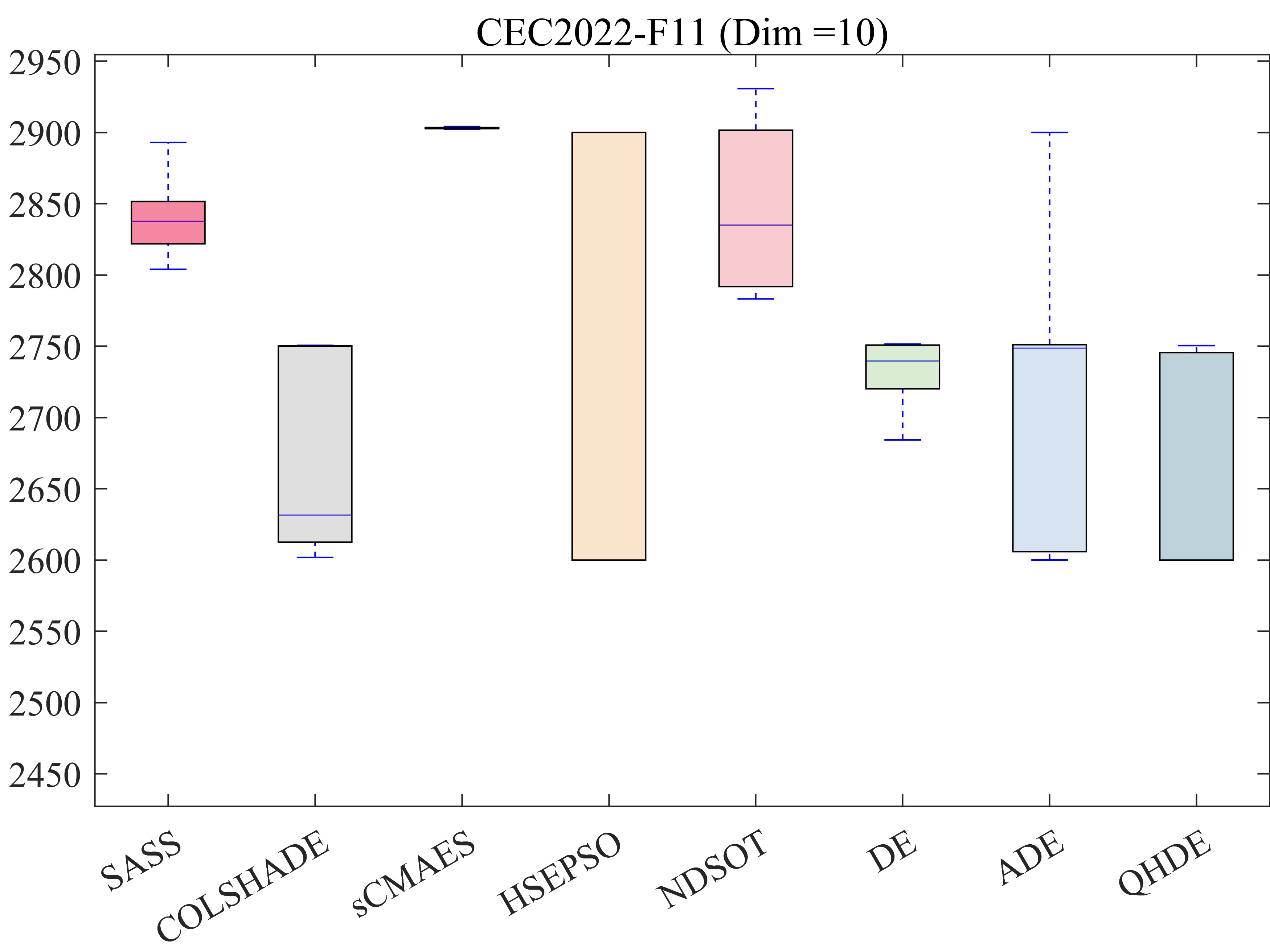}

	\vspace{0.8em}
	
	\includegraphics[width=0.32\linewidth]{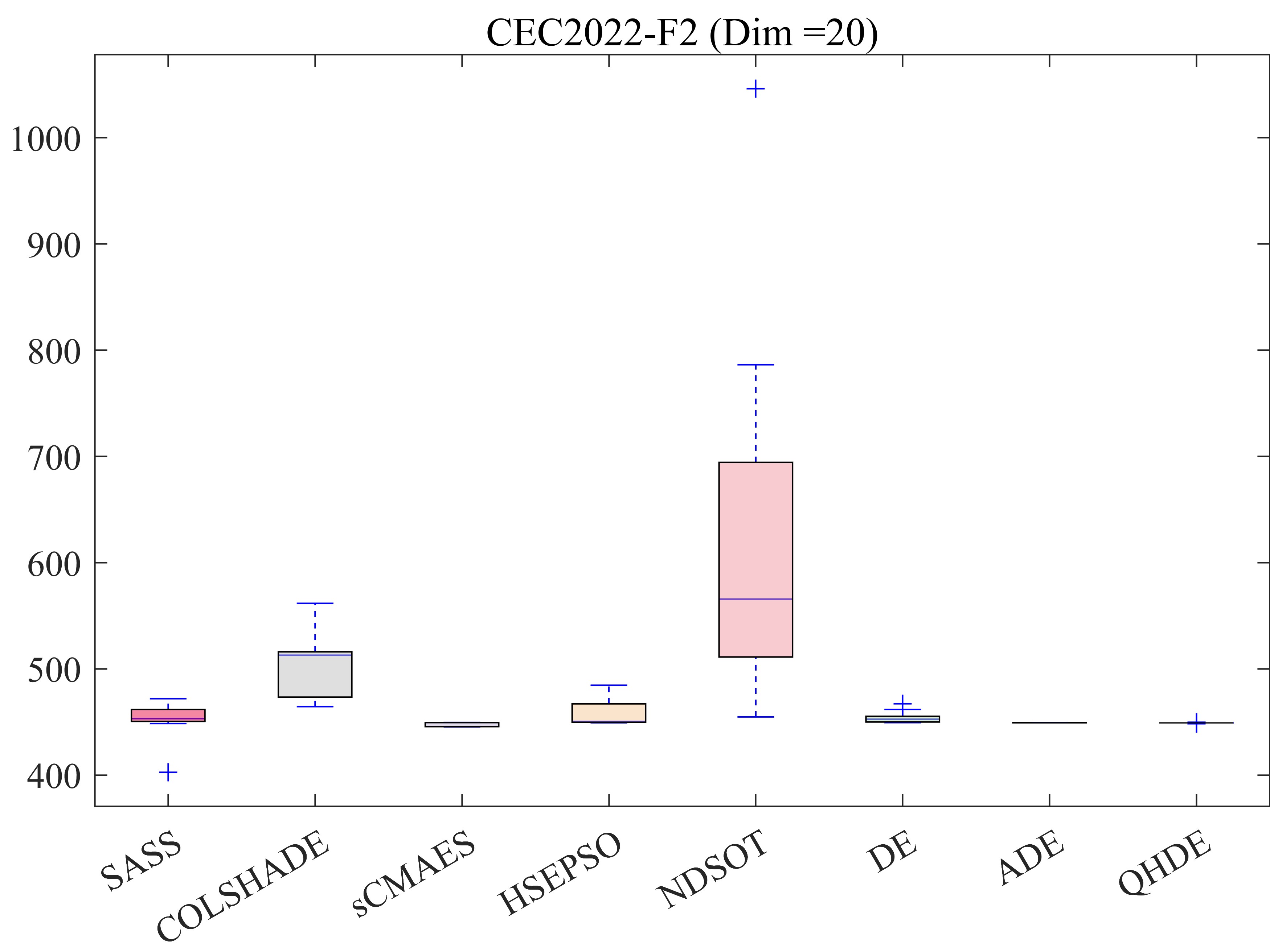}%
	\hfill
	\includegraphics[width=0.32\linewidth]{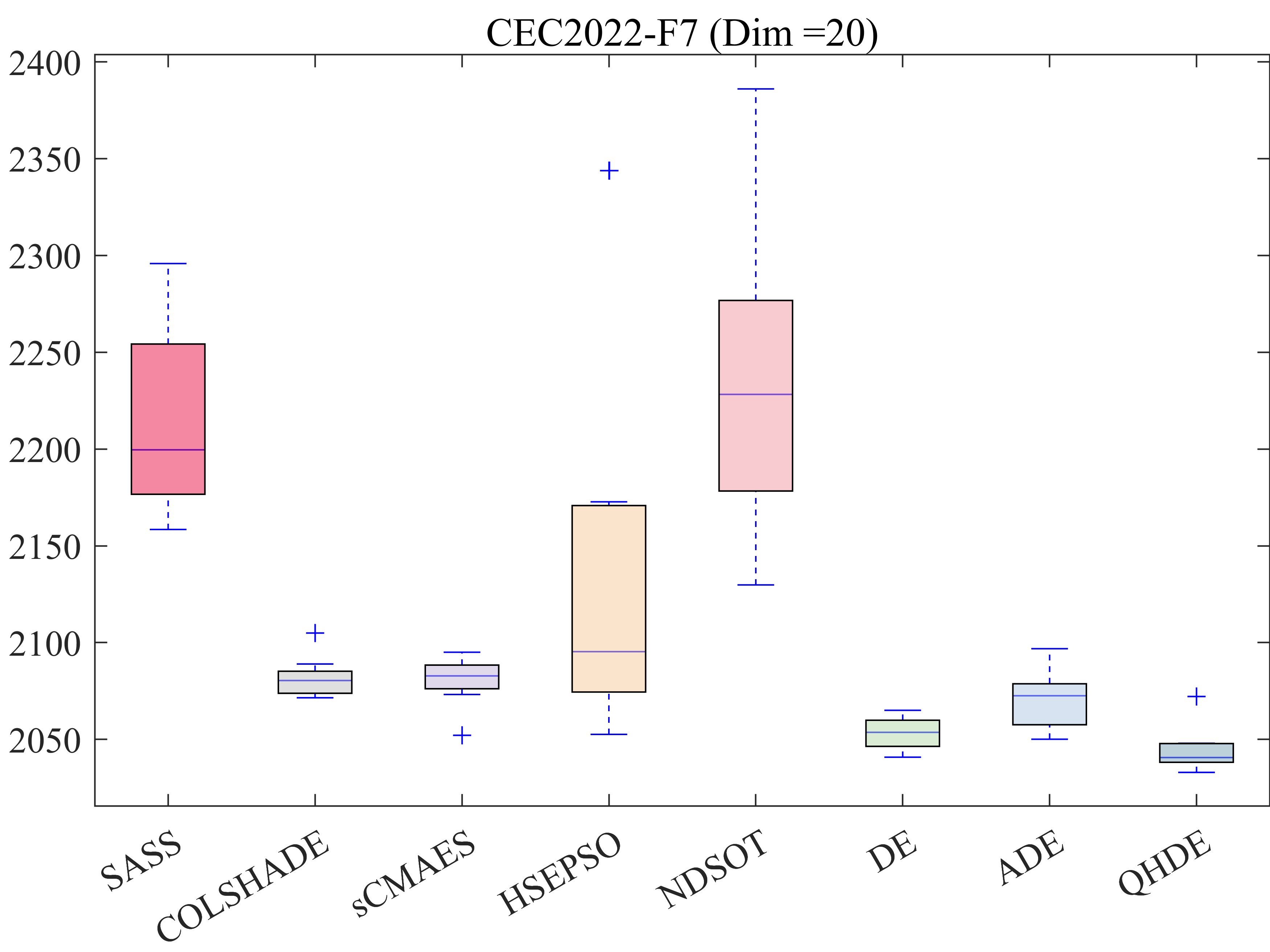}%
	\hfill
	\includegraphics[width=0.32\linewidth]{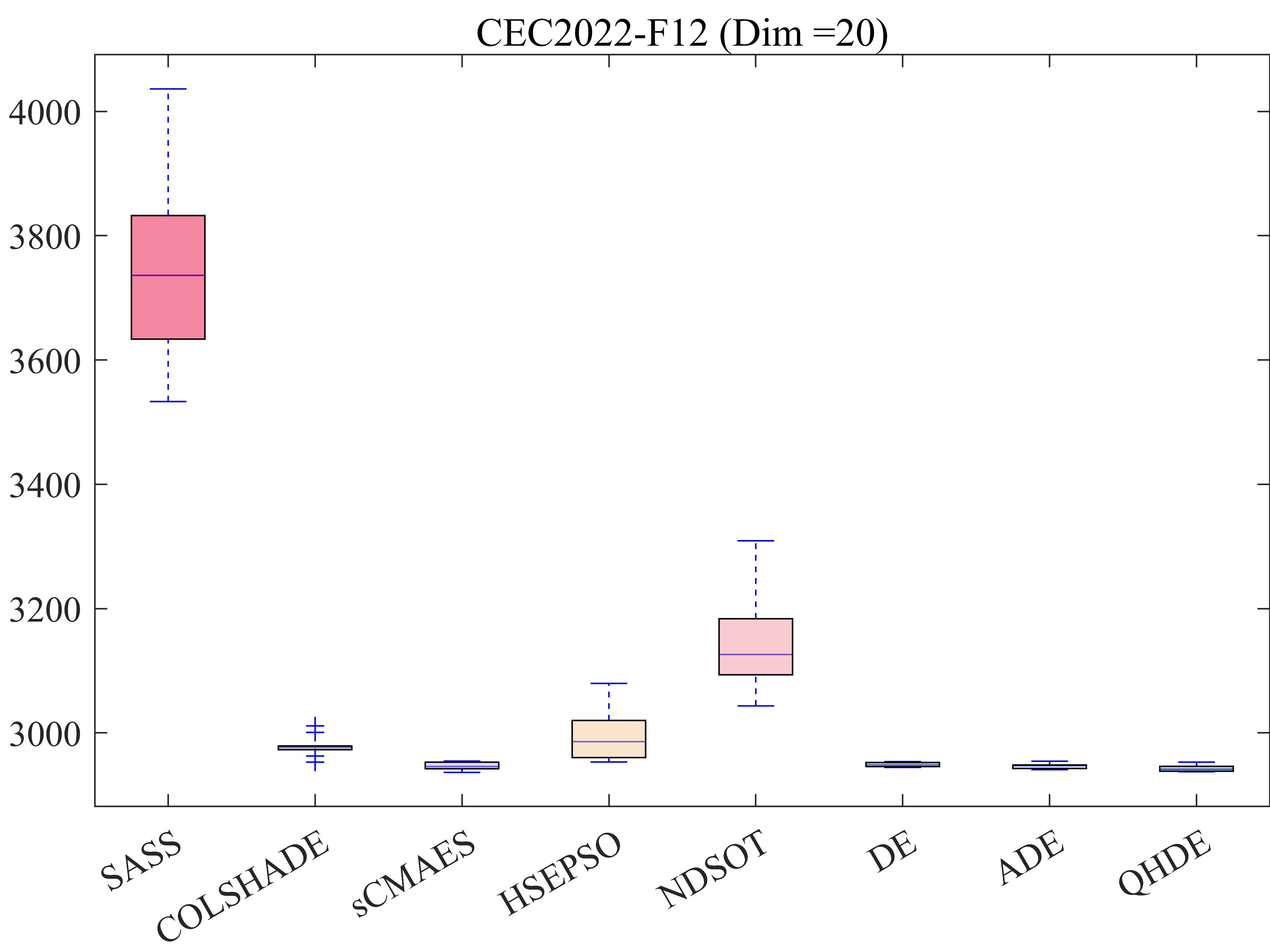}
	
	\caption{Box plots of QHDE and other algorithms on CEC 2022 functions.}
	\label{fig:box}
\end{figure}

As shown in Figure~\ref{fig:box}, the boxplot analysis of various algorithms on representative functions reveals a significant stability advantage of QHDE. The box of the QHDE boxplot is relatively narrow, indicating a highly concentrated data distribution. This suggests that, across multiple experimental runs, the results generated by QHDE exhibit minimal dispersion, and the search process remains stable. By leveraging its three integrated mechanisms, QHDE effectively avoids local optima to provide stable, high-quality solutions.
In contrast, for algorithms such as SASS, HSEPSO and NDSOT, the overall positions of the boxplots are relatively higher. This suggests that their results are relatively suboptimal and that their robustness is weaker.

\subsection{Ablation experiments}
\begin{figure}[t]
	\centering
	\includegraphics[width=0.9\linewidth]{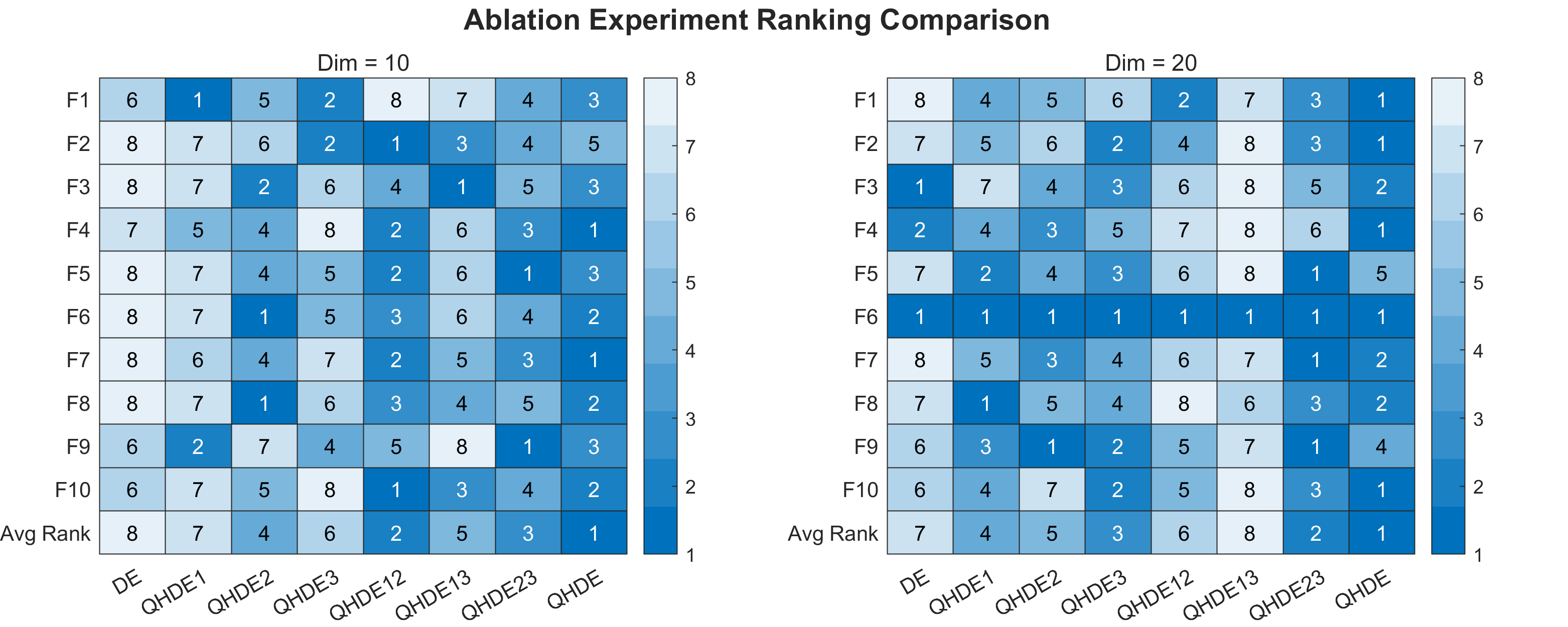}
	\caption{Ablation experiment on CEC 2020.}
	\label{fig:ablation}
\end{figure}

QHDE incorporates three enhancement strategies: the good point set-chaos reverse learning initialization mechanism (Strategy 1), the potential-driven dynamic quantum tunneling strategy (Strategy 2), and the dynamic elite pool combined with Cauchy-Gaussian mixed perturbation (Strategy 3). To examine the individual and combined effects of these strategies on QHDE'Ds performance, ablation experiments were conducted. We designed six variants (QHDE1, QHDE2, QHDE3, QHDE12, QHDE13, and QHDE23) to isolate these components. Figure~\ref{fig:ablation} display the Friedman rankings on CEC 2020 (Dim=10, 20), among which the improvement of QHDE is the most remarkable.

 Figure~\ref{fig:ablation} demonstrate that QHDE achieves the highest average Friedman ranking, consistently placing in the top 5 across all test functions. Among the three strategies, Strategy 2 (quantum tunneling) serves as the primary performance driver, as variants incorporating it (QHDE2, 12, 23) consistently outrank others. While Strategy 1 (initialization) offers limited independent gains, Strategy 3 (mixed perturbation) shows increased efficacy in high-dimensional scenarios. Its synergy with Strategy 2 enables QHDE and QHDE23 to maintain superior optimization stability and global search capability as problem complexity scales.

\subsection{Portfolio selection}
This section evaluates QHDE across four portfolio selection problems (comprising 20, 40, 60, and 80 stocks) to verify its optimization superiority. The dataset includes closing prices from the CSI 300 Index (Shanghai and Shenzhen Stock Exchanges) between January 1, 2024, and December 31, 2024. Stocks were selected based on their average daily returns to form the test portfolios. To ensure a fair comparison, all eight algorithms utilized a population size of 50 over 100 iterations. Performance was measured using three key metrics: the objective function F(E), the Sharpe ratio Sr, and the equality constraint S(E). The results are shown in Table~\ref{tab:portfolio_results}, with the best performers highlighted in bold.

\begin{table*}[t]
	\caption{Four portfolios' results solved by seven metaheuristic algorithms}
	\label{tab:portfolio_results}
	\centering
	\resizebox{\textwidth}{!}{
	\begin{tabular}{lccccccccccccccc}
		\toprule
		\multirow{2}{*}{Algorithm} &
		\multicolumn{3}{c}{20 stacks} & 
		\multicolumn{3}{c}{40 stacks} &
		\multicolumn{3}{c}{60 stacks} &
		\multicolumn{3}{c}{80 stacks} \\
		\cmidrule(lr){2-4}
		\cmidrule(lr){5-7}
		\cmidrule(lr){8-10}
		\cmidrule(lr){11-13}
		& F(E)  & Sr & S(E) 
		& F(E)  & Sr & S(E)
		& F(E)  & Sr & S(E)
		& F(E)  & Sr & S(E) \\
		\midrule
		SASS    & -2.37E+04	&4.26E+00&	1.15E+00
		&	-6.49E+05&	3.49E+00&	1.81E+00
		&	-5.36E+06&	2.42E+00&	3.32E+00
		&	-1.43E+07&	1.58E+00&	4.79E+00 \\

		COLSHADE & 1.59E+01&	1.60E+01&	1.00E+00
		&	-5.27E+05&	6.40E+00&	1.73E+00
		&	-1.59E+07&	3.19E+00&	4.99E+00
		&	-1.57E+08&	6.14E-01&	1.35E+01
		 \\
		
		sCMAES  & 1.93E+01&	2.02E+01&	1.00E+00
		&	-6.66E+06&	3.10E+00&	3.58E+00
		&	-1.67E+07&	1.66E+00&	5.08E+00
		&	-8.43E+07&	8.17E-01&	1.02E+01
		 \\
		
		HSEPSO  &1.25E+01&	1.25E+01&	10.00E-01
		&	1.12E+01&	1.12E+01&	1.00E+00
		&	-1.67E+05&	5.57E+00&	1.41E+00
		&	-1.80E+06&	3.00E+00&	2.34E+00 
		 \\
		
		NDSOT   & 9.00E+00&	9.00E+00&	1.00E+00
		&	-1.60E+07&	7.14E-01&	5.00E+00
		&	-2.89E+08&	4.89E-01&	1.80E+01
		&	-5.29E+08&	4.42E-01&	2.40E+01
		 \\
		
		DE      & 1.36E+01&	1.45E+01&	9.99E-01
		&	6.87E+00&	7.09E+00&	1.00E+00
		&	-1.51E+06&	3.62E+00&	2.23E+00
		&	-2.14E+07&	1.29E+00&	5.63E+00
		
		 \\
		
		ADE     & 8.19E+00&	1.04E+01&	9.98E-01
		&	7.74E+00&	7.88E+00&	1.00E+00
		&	8.53E+00&	8.53E+00&	1.00E+00
		&	3.03E+00&	3.07E+00&	10.00E-01
		
		 \\
		
		QHDE    & \textbf{2.81E+01} & \textbf{2.87E+01} & 1.00E+00 
		& \textbf{1.63E+01} & \textbf{1.71E+01} & 1.00E+00 
		& \textbf{1.23E+01} & \textbf{1.37E+01} & 9.99E-01 
		& \textbf{5.96E+00} & \textbf{1.02E+01} & 1.00E+00 \\
		\bottomrule
	\end{tabular}
}
\end{table*}

Table~\ref{tab:portfolio_results} demonstrates that QHDE maintains remarkable and stable performance across all portfolio scales. Its objective values ($F(E)$) of 28.1, 16.3, 12.3, and 5.96 are the highest in their respective categories, outperforming the runners-up by margins of 45.5\%, 45.5\%, 44.1\%, and 96.6\%. Furthermore, the Sr of QHDE consistently surpasses that of other comparative algorithms, indicating its effectiveness in balancing risk and return to achieve superior risk-adjusted returns.
It is worth noting that in small-scale tasks, the optimal value of DE ranks in the second tier, suggesting that DE possesses a certain degree of exploitation capability in complex portfolio problems. However, as the scale increases, the performance of DE deteriorates rapidly. Specifically, its S(E) value at 60 dimensions is significantly higher than expected, revealing its limitations in handling complex constraints. In contrast, the proposed QHDE maintains values extremely close to 1 across all test scales. It not only improves optimization precision but also effectively resolves the deficiencies of DE in constraint handling, thereby exhibiting superior comprehensive performance.

\begin{figure}[t]
	\centering
	\subfloat[]{\includegraphics[width=0.32\linewidth]{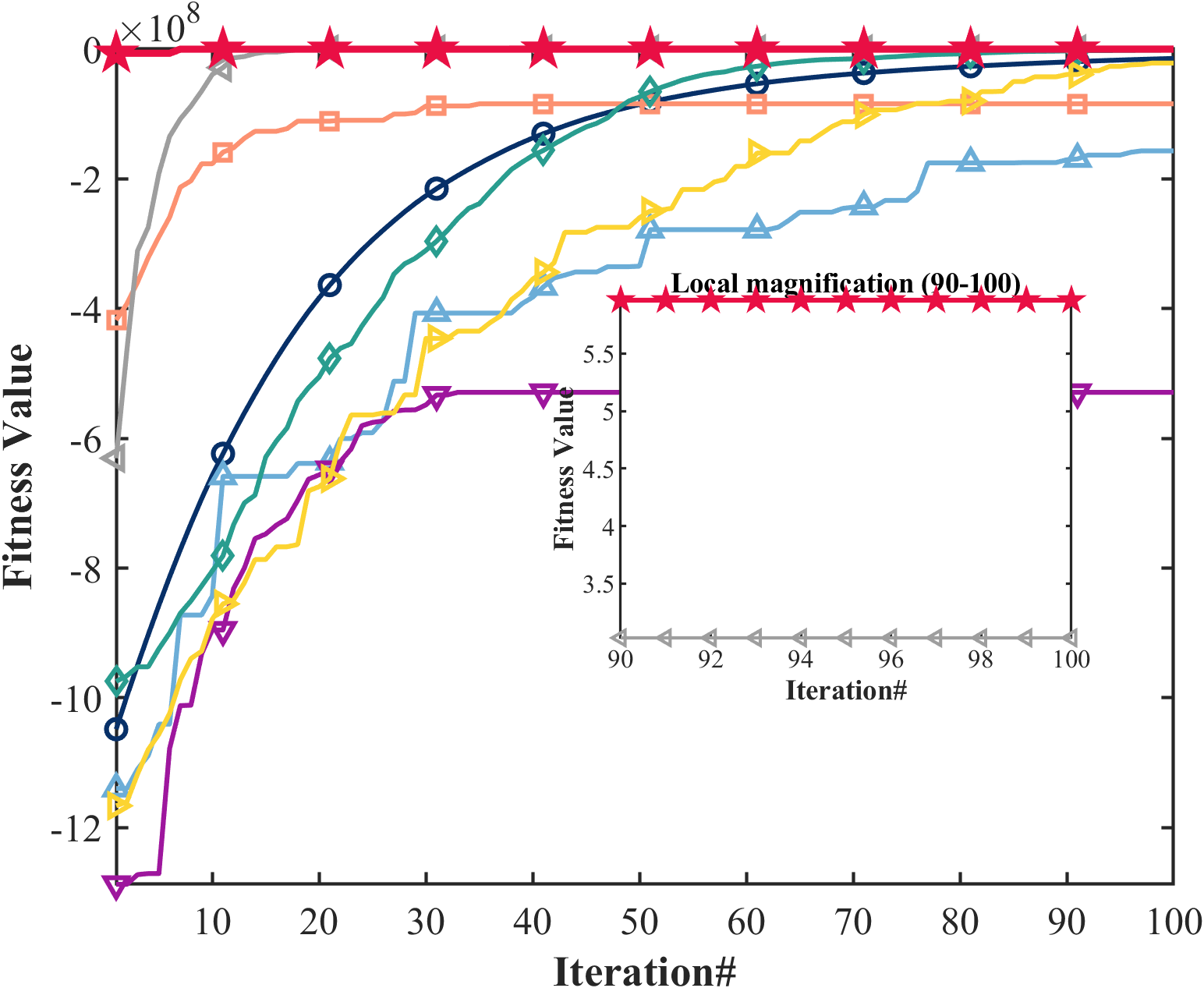}}
	\hfill
	\subfloat[]{\includegraphics[width=0.32\linewidth]{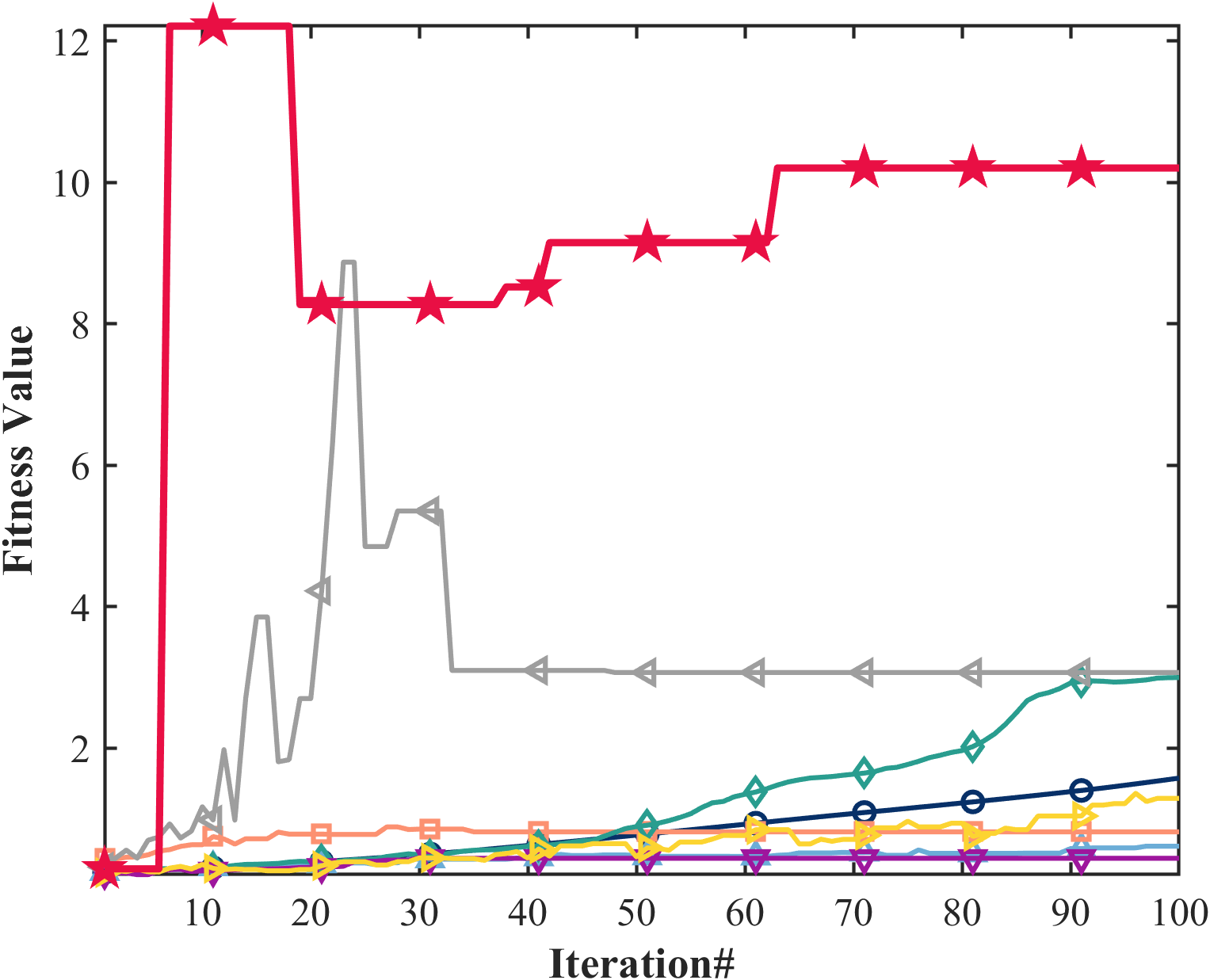}}
	\hfill
	\subfloat[]{\includegraphics[width=0.32\linewidth]{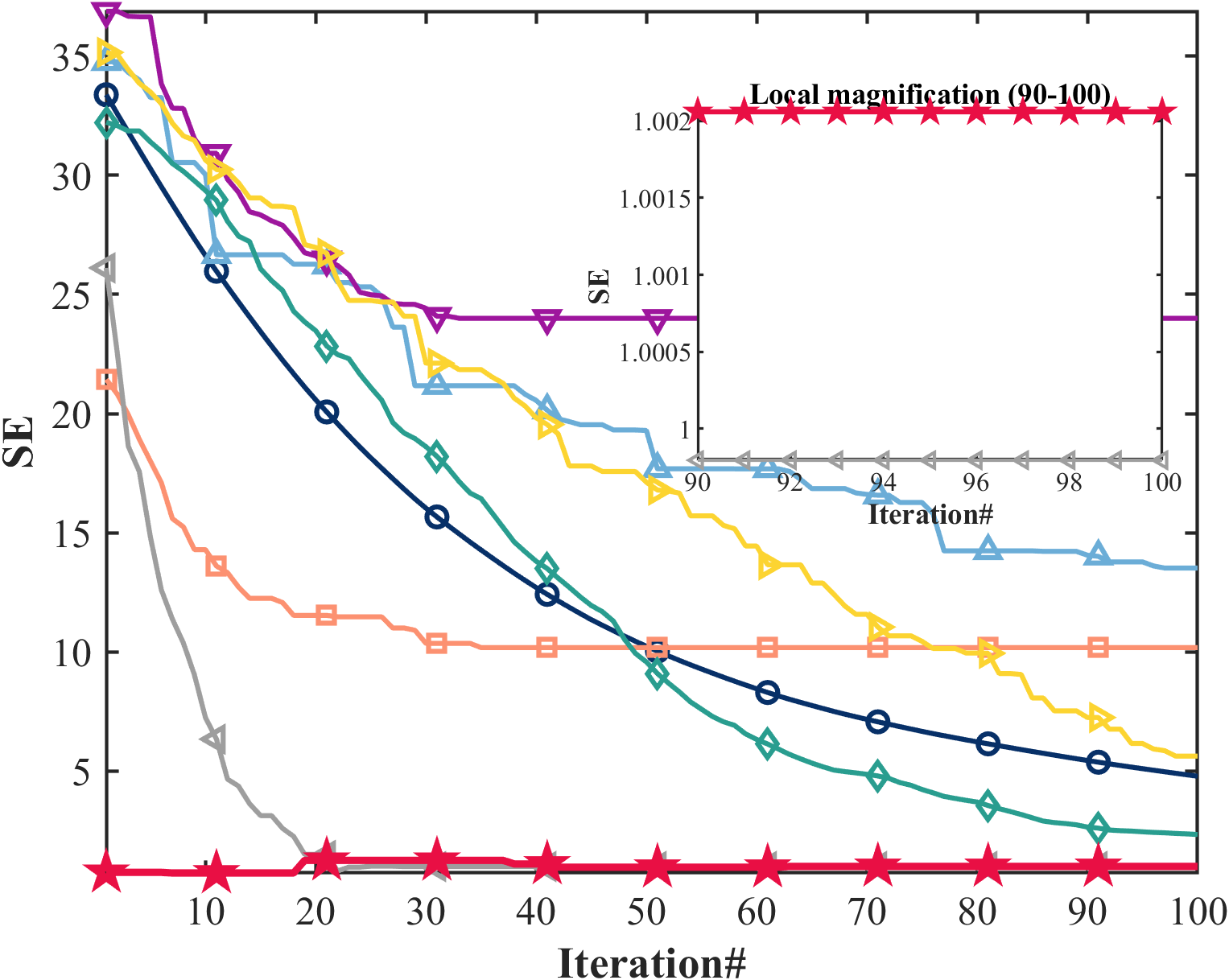}}
	
	\vspace{3pt}
	
	\includegraphics[width=0.4\linewidth]{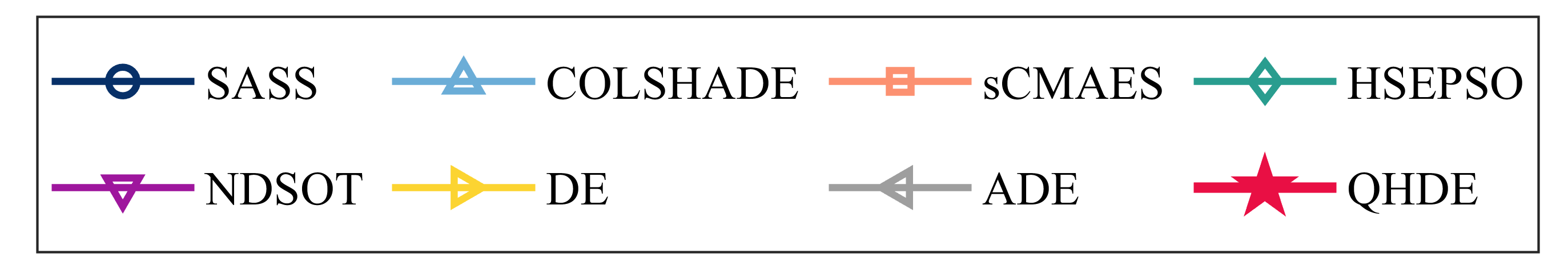}
	
	\caption{Evaluation and comparison of QHDE applied to 80 stocks alongside other metaheuristic algorithms. (a) illustrates the objective function value F(E), (b) presents the Sharpe ratio Sr, and (c) presents the equality constraint S(E).}
	\label{fig:abc}
\end{figure}

Figure \ref{fig:abc} presents the iterative process for 80 stocks (see Appendix \cite{QHDE} for other scales). 
As shown in Figure~\ref{fig:abc}(a), QHDE achieved the highest F(E) value and had the fastest convergence speed, highlighting its significant advantages in global search and convergence efficiency. However, other algorithms like DE failed to converge to the optimal solution after 100 iterations, further demonstrating their limitations in high-dimensional optimization problems.
In Figure~\ref{fig:abc}(b), although ADE converges slightly faster initially, QHDE ultimately yields a superior Sr. This indicates that QHDE is capable of consistently providing high returns while controlling risk, demonstrating good risk-return balance.
Crucially, Figure~\ref{fig:abc}(c) highlights QHDE's reliability in constraint handling. It rapidly approaching 1 during the initial iterations, competitors like sCMAES and NDSOT struggle or fail to converge. The magnified view confirms that QHDE maintains an error-free, stable state throughout the optimization process.
 
In summary, QHDE demonstrates a significant advantage in complex portfolio optimization. It substantially enhances both global exploration and local exploitation, while maintaining exceptional adaptability and reliability when addressing high-dimensional, constrained problems. These characteristics establish QHDE as a robust and precise tool for solving complex financial optimization tasks and portfolio selection problems.
\section{Conclusion}
In this paper, we propose a quantum-enhanced DE algorithm for financial portfolio optimization. QHDE integrates three major strategies: (1) good-point-set chaos-reverse initialization to boost population diversity; (2) a potential-driven dynamic quantum tunneling strategy that enables individuals to bypass classical search barriers; and (3) a dynamic elite pool with Cauchy-Gaussian perturbation to prevent premature convergence through strategic late-stage disturbances. Together, these mechanisms accelerate global convergence and improve robustness in complex search spaces.

The effectiveness of QHDE is demonstrated on the CEC 2020 and CEC 2022 benchmark suites, where it is compared with seven state-of-the-art optimizers. Statistical results show that QHDE consistently outperforms the competitors. In portfolio optimization experiments, QHDE achieves superior returns, lower risk, and faster convergence, exhibiting strong global exploration and local exploitation capabilities. These advantages position QHDE as a competitive solver for practical financial optimization tasks.

Future research could extend the application of QHDE to other financial optimization challenges, such as multi-objective investment strategies and dynamic portfolio management. Furthermore, comprehensive studies on the key parameters of quantum tunneling strategies could further refine the algorithm's performance.

\begin{credits}
\subsubsection{\ackname} This work was supported by Tianjin Science and Technology Major Project (24ZXZSSS00420), Major Program of the National Natural-Science Foundation of China (62233011).

\subsubsection{\discintname} The authors declare that they have no conflicts of interest.
 
\end{credits}
%
%
%
\bibliographystyle{splncs04}
\bibliography{sample-base}
\end{document}